\documentclass{SCIS2025}
\usepackage{array}
\usepackage{textcomp}
\usepackage{stfloats}
\usepackage{verbatim}
\usepackage{graphicx}
\usepackage{caption}
\usepackage{subcaption}
\usepackage{wrapfig} 

\usepackage{cite}
\usepackage{dsfont}
\usepackage{algorithm}
\usepackage{algorithmic}
\usepackage{xcolor}
\definecolor{bluejoint}{HTML}{0b0bff}
\definecolor{orangejoint}{HTML}{fca200}
\usepackage[table]{xcolor}
\usepackage{pifont}
\usepackage{multirow}
\usepackage{bm}
\definecolor{myColor}{RGB}{26, 153, 230}
\newcommand{\bl}[1]{\textcolor{myColor}{#1}}
\usepackage{booktabs}
\usepackage{enumitem}
\usepackage{footnote}
\definecolor{cvprblue}{rgb}{0.21,0.49,0.74}
\usepackage[capitalize,nameinlink,noabbrev]{cleveref}

\crefname{section}{Sec.}{Secs.}
\Crefname{section}{Section}{Sections}
\Crefname{table}{Table}{Tables}
\crefname{table}{Tab.}{Tabs.}

\usepackage{tabularx}
\newcolumntype{A}{>{\centering\arraybackslash}X} 
\usepackage{threeparttable}

\colorlet{cyan}{cyan!10}

\usepackage{amssymb, bm}
\usepackage{mathtools}   
\usepackage{xspace}

\DeclarePairedDelimiter{\paren}{\lparen}{\rparen}       
\DeclarePairedDelimiter{\bracket}{\lbrack}{\rbrack}     
\DeclarePairedDelimiter{\norm}{\lVert}{\rVert}          
\newcommand{\M}[1]{\mathbb{M}\!\bracket*{#1}}
\newcommand{\KL}[2]{D_{\mathrm{KL}}\!\paren{#1\,\|\,#2}}

\begin{document}
\ArticleType{RESEARCH PAPER}
\Year{2025}
\Month{January}
\Vol{68}
\No{1}
\DOI{}
\ArtNo{}
\ReceiveDate{}
\ReviseDate{}
\AcceptDate{}
\OnlineDate{}
\AuthorMark{}
\AuthorCitation{}

\title{MMCL: Correcting Content Query Distributions for Improved Anti-Overlapping X-Ray Object Detection}{MMCL: Correcting Content Query Distributions for Improved Anti-Overlapping X-Ray Object Detection}


\author[1,2]{Tong Jia}{}
\author[1,3]{Mingyuan Li}{2310289@stu.neu.edu.cn}
\author[1]{Hui Lu}{}
\author[1]{Hao Wang}{}
\author[1]{Bowen Ma}{}
\author[1]{Shiyi Guo}{}
\author[1]{\\Shuyang Lin}{}
\author[1]{Dongyue Chen}{}
\author[1]{Haoran Wang}{}
\author[3]{Baosheng Yu}{}


\address[1]{College of Information Science and Engineering, Northeastern University, Shenyang, 110819, China}
\address[2]{State Key Laboratory of Synthetical Automation for Process Industries, Northeastern University, Shenyang, 110819, China}
\address[3]{Lee Kong Chian School of Medicine, Nanyang Technological University, 639798, Singapore}

\abstract{Unlike natural images with occlusion-based overlap, X-ray images exhibit depth-induced superimposition and semi-transparent appearances, where objects at different depths overlap and their features blend together. These characteristics demand specialized mechanisms to disentangle mixed representations between target objects (e.g., prohibited items) and irrelevant backgrounds. While recent studies have explored adapting detection transformers (DETR) for anti-overlapping object detection, the importance of well-distributed content queries that represent object hypotheses remains underexplored. 
In this paper, we introduce a multi-class min-margin contrastive learning (MMCL) framework to correct the distribution of content queries, achieving balanced intra-class diversity and inter-class separability. The framework first groups content queries by object category and then applies two proposed complementary loss components: a multi-class exclusion loss to enhance inter-class separability, and a min-margin clustering loss to encourage intra-class diversity.
We evaluate the proposed method on three widely used X-ray prohibited-item detection datasets, PIXray, OPIXray, and PIDray, using two backbone networks and four DETR variants. Experimental results demonstrate that MMCL effectively enhances anti-overlapping object detection and achieves state-of-the-art performance on both datasets. Code will be made publicly available on GitHub.}


\maketitle

\section{Introduction}
Anti-overlapping object detection is a key challenge in X-ray image understanding, where objects at different degrees overlap, producing mixed, semi-transparent, and coupled signals. Unlike natural images with opaque occlusions, these overlaps distort object boundaries, textures, and color, hindering accurate recognition~\cite{mcketty1998aapm, swinehart1962beer, wang2017chestx}. The problem becomes more pronounced in security inspection scenes~\cite{SIXray, OPIXray, tao2021towards}, where luggage is often crowded with tightly packed items, leading to severe feature overlapping. Detecting prohibited objects, such as knives, guns, or explosives, in complex conditions is essential to ensuring public safety. Advanced anti-overlapping detection methods enable models to disentangle superimposed features~\cite{AO-DETR, SIXray}, uncover concealed threats~\cite{PIDray,CLCXray}, and reduce false alarms~\cite{GADet}, thereby enhancing the accuracy and reliability of automated screening systems deployed in airports, customs, and other security environments.

\begin{figure}[t]
\centering
\includegraphics[width=0.6\linewidth]{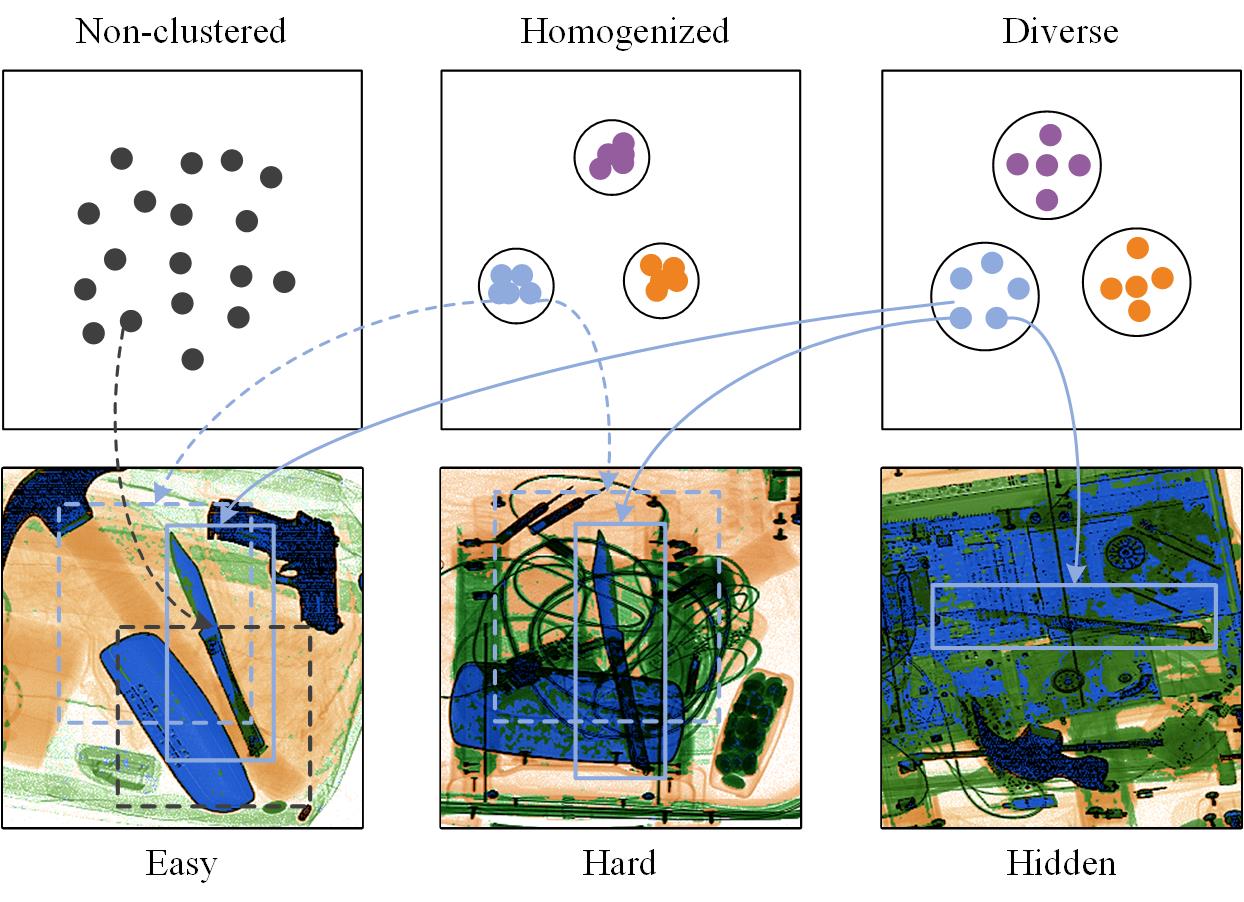}
\caption{Illustration of different content query distributions for anti-overlapping object detection.
\textbf{Left:} Non-clustered content queries~\cite{Deformable-DETR,RT-DETR,DINO} can recognize prohibited items (e.g., knives) only in simple overlapping scenes (Easy).
\textbf{Middle:} Content queries with intra-class compactness~\cite{AO-DETR} become homogenized and can handle moderately complex overlapping scenes (Hard). \textbf{Right:} Content queries that maintain intra-class diversity and inter-class separability can effectively address complex and heavily overlapping scenes (Hidden).}
\label{fig:fig1}
\end{figure}

Following the success of general object detection in natural images~\cite{Faster, Mask}, particularly the detection transformer or DETR~\cite{DETR} and its variants~\cite{Deformable-DETR, DINO, DAB-DETR, RT-DETR}, recent studies have adapted these models for prohibited item detection in X-ray images. These adaptations introduce specialized anti-overlapping mechanisms, such as adaptive label assignment~\cite{GADet, CLCXray, Xdet} and attention-based background suppression~\cite{PIDray, CLCXray, OPIXray, SIXray, FDTNet}, to address the challenges posed by crowded and overlapping objects. However, the influence of different content query distributions for anti-overlapping object detection remains underexplored. As shown in~\cref{fig:fig1}, non-clustered content queries detect prohibited items (e.g., knives) only in simple overlapping scenes~\cite{Deformable-DETR, RT-DETR, DINO}. Queries with intra-class compactness~\cite{AO-DETR} generalize across categories but struggle in varied backgrounds. In contrast, well-distributed content queries that maintain intra-class diversity and inter-class separability perform effectively in complex and heavily overlapping scenarios. Therefore, correcting content query distributions in DETR-like detectors to balance intra- and inter-class relationships is essential for improving anti-overlapping performance in X-ray object detection.

To address this issue, contrastive learning provides a practical framework for learning discriminative embeddings by pulling semantically similar samples together and pushing dissimilar ones apart. In DETR-like detectors, it can help correct content query distributions, enhancing inter-class separability while preserving intra-class coherence for overlapping objects. A explicit and common approach partitions queries into class-specific groups~\cite{AO-DETR} and applies a contrastive loss, such as N-pair~\cite{N-pair}, InfoNCE~\cite{InfoNCE}, IIC~\cite{IIC_loss}, ICE~\cite{ICE_loss}, or OCA~\cite{OCA_loss}. However, these conventional losses often overemphasize inter-class separation, leading to overly compact intra-class clusters that fail to capture subtle variations among overlapping instances. Addressing this requires a more flexible mechanism that maintains intra-class diversity while balancing inter-class separability for robust anti-overlapping detection.

\begin{figure}[t]
\centering
\includegraphics[width=0.7\linewidth]{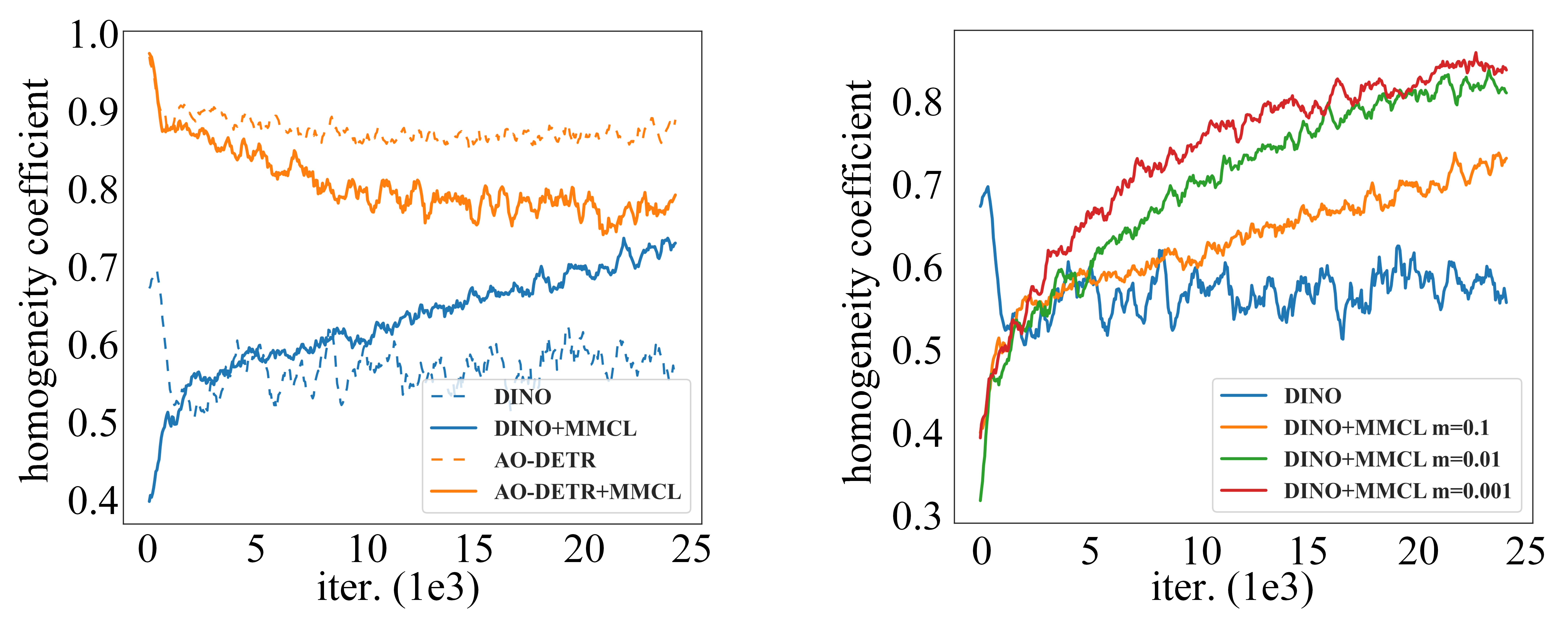}
\caption{Illustration of intra-class diversity in content queries with and without MMCL. \textbf{Left:} MMCL enhances the homogeneity coefficient in DINO (non-clustered) while reducing it in AO-DETR (over-compact). \textbf{Right:} The hyperparameter $m$ in the proposed loss function enables flexible control over intra-class margins. The homogeneity coefficient is the average cosine similarity among intra-class queries; a higher value indicates greater homogeneity and lower diversity.}
\label{fig:fig2}
\end{figure}

In this paper, we propose Multi-class Min-margin Contrastive Learning (MMCL) to correct content query distributions for improved anti-overlapping X-ray object detection. MMCL partitions queries into category-specific groups and refines their distributions via a novel contrastive loss with two components: an Inter-class Moderate Exclusion (IME) loss to enforce inter-class separability and an Intra-class Min-margin Clustering (IMC) loss to preserve intra-class diversity, with a hyperparameter $m$ controlling the minimum intra-class margin. As shown in \cref{fig:fig2}, MMCL effectively corrects non-clustered distributions in DINO~\cite{DINO} and over-compact distributions in AO-DETR~\cite{AO-DETR}, while $m$ allows flexible control over intra-class margins.

We evaluate MMCL on three widely used X-ray prohibited item detection datasets, PIXray~\cite{PIXray}, OPIXray~\cite{OPIXray}, and PIDray~\cite{PIDray}. Experiments across four DETR variants and two backbones show that MMCL consistently improves anti-overlapping detection by refining content query distributions. On PIXray, it boosts RT-DETR~\cite{RT-DETR} from 62.3\% to 63.6\% AP, DINO~\cite{DINO} from 64.3\% to 66.7\% AP, and AO-DETR~\cite{AO-DETR} from 73.9\% to 74.6\% AP, among others. Similar gains are observed on OPIXray. Quantitative analysis on PIDray demonstrates that MMCL is particularly effective for severe overlapping scenes. Notably, integrating MMCL with AO-DETR (Swin-L) achieves new state-of-the-art performance on OPIXray, demonstrating its effectiveness as a plug-and-play module for enhancing anti-overlapping detection.

The main contributions of this paper are as follows:
\begin{enumerate}
\item We demonstrate that correcting content query distributions in DETR variants is essential for enhancing anti-overlapping X-ray object detection.
\item We develop a unified framework, MMCL, that employs contrastive learning to effectively correct content query distributions in DETR-based detectors.
\item We introduce a specialized contrastive loss that adaptively balances intra-class diversity and inter-class separability, enabling more discriminative content query representations.
\end{enumerate}

\section{Related Work}\label{Related Work}

This section reviews advances in three areas: prohibited-item detection, the key challenge of overlapping X-ray objects; DETR variants, which underpin modern object detection; and contrastive learning, used to learn representations by modeling pairwise sample relationships.

\subsection{Prohibited Item Detection}
Prohibited item detection~\cite{OPIXray, PIDray, AO-DETR} identifies overlapping objects in X-ray images, where transparency complicates recognition. Existing methods include label assignment strategies~\cite{GADet, CLCXray, Xdet} for improved localization and attention-based mechanisms~\cite{PIDray, CLCXray, OPIXray, SIXray, FDTNet, DvXray} to extract foreground features while suppressing background interference. Most rely on CNNs~\cite{CNN, ResNet, wang2025exploring, chen2024augmentation, zhang2024comprehensive}, which lag behind transformer-based detectors~\cite{Swin-Transformer, CO-DETR, zhang2025comprompter} and struggle to link classification and localization~\cite{GADet}. While GADet~\cite{GADet} leverages geometric priors for stable intra-class and discriminative inter-class features, AO-DETR~\cite{AO-DETR} embeds category semantics into queries to better extract overlapping features. Despite these advances, AO-DETR remains complex and inflexible, motivating further exploration of transformer-based detectors and contrastive learning for robust anti-overlapping X-ray detection.

\subsection{Detection Transformers}
Since the introduction of DETR, transformer-based detectors have rapidly advanced beyond CNN-based models. Its query-based one-to-one label assignment removes the need for anchors and NMS, but convergence remains slow. Two main strategies address this: one combines one-to-many supervision from conventional detectors~\cite{Faster, ATSS, YOLOX} with one-to-one DETR~\cite{H-DETR, DAC-DETR, CO-DETR, Group-DETR, Stable-DETR, zhang2025mr}, while the other enriches queries with localization~\cite{Deformable-DETR}, anchor~\cite{Anchor-DETR, DAB-DETR}, or denoising information~\cite{DN-DETR, DINO}. However, these methods generally lack explicit category information. AO-DETR~\cite{AO-DETR} addresses this by embedding class semantics into queries via a query-specific label assignment, improving anti-overlapping detection. Inspired by this, we propose MMCL, a more flexible mechanism to refine content query distributions for overlapping object detection.

\subsection{Contrastive Learning}

Contrastive learning learns discriminative representations by pulling together samples from the same class and pushing apart samples from different classes. It can be broadly divided into self-supervised and supervised approaches. Self-supervised methods include instance-wise contrastive learning (e.g., SimCLR~\cite{SimCLR}, MoCo~\cite{MoCo}), which treats augmented views of the same instance as positive pairs, and cluster-based learning~\cite{CC, TCL}, which generates pseudo-labels via clustering for supervised contrastive training. Loss functions can categorize supervised contrastive learning into softmax-based~\cite{N-pair, InfoNCE, Circle_loss, SupCon_loss, OCA_loss}, cross-entropy-based~\cite{EBM, C2AM}, and decoupled designs~\cite{C2AM, ICE_loss, IIC_loss} that separately optimize intra-class attraction and inter-class repulsion. The decoupled paradigm is particularly flexible, balancing intra-class diversity and inter-class separability, yet existing methods generally lack mechanisms to preserve intra-class diversity, which is crucial for tasks such as refining content query distributions in anti-overlapping DETR-based detection.

\begin{figure*}[t]
    \centering
    \includegraphics[width=\linewidth]{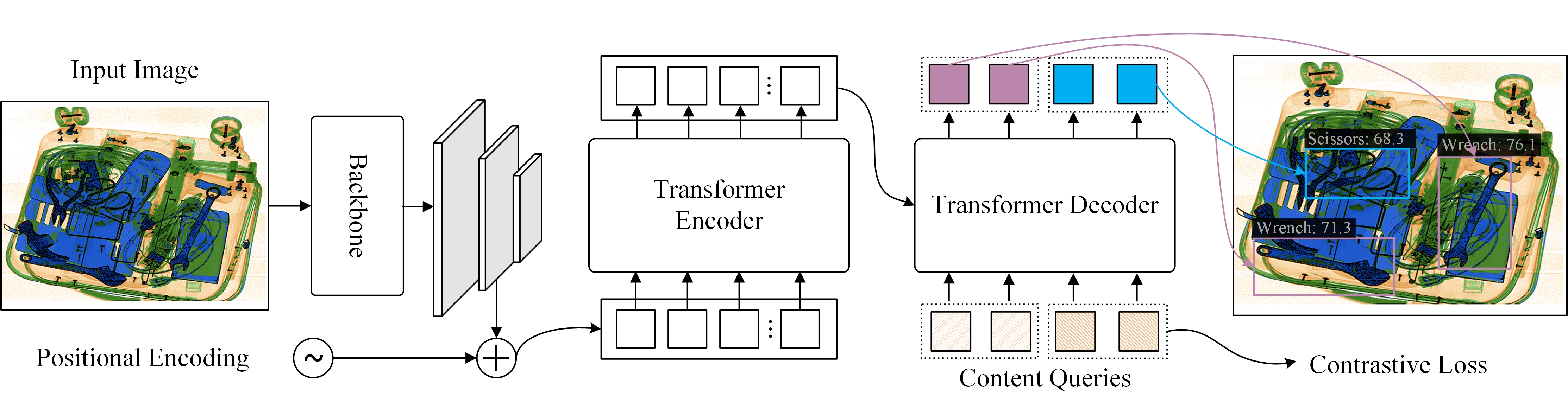}
    \caption{Overview of the proposed MMCL framework for anti-overlapping X-ray object detection. The framework integrates a contrastive loss to refine the distribution of content queries, thereby enhancing object discrimination and reducing overlap confusion—all without modifying the underlying architecture.}
    \label{fig:fig3}
\end{figure*}

\begin{wrapfigure}{r}{0.5\linewidth}
\vspace{-20pt} 
\begin{minipage}{\linewidth} 
\begin{algorithm}[H]
\caption{Summary of key steps in MMCL.}\label{Algorithm_MMCL}
\begin{algorithmic}
\REQUIRE Let $K$ denote the number of categories, $L$ the set of all decoder layers, $T$ the target layers, $P$ the prediction results of all decoder layers, $G$ the set of ground-truth objects, $\mathcal{H}$ the inherent Hungarian label assignment strategy, and $\mathbf{Q}$ the content queries of all decoder layers.
\ENSURE Learned content queries $\mathbf{Q}$.\\ 
\STATE \textbf{Forward pass:}
\STATE Initialize the total loss $\mathcal{L}$ to 0;
\FOR{ $\forall$ decoder layer index $ l\in L$ }
\STATE $\{ P_i^l; G_i \}\gets \mathcal{H}^l(P^l,G)$;
\STATE $\mathcal{L}^l \gets$ $\mathcal{L}_{\text{base}}(\{P_i^l; G_i \})$;
\IF{$l \in T$}
\STATE Divide $\mathbf{Q}^l$ into $\{\mathbf{Q}_{k}^l\}_{k=1}^K$ by $K$;
\STATE $\mathcal{L}^l \gets \mathcal{L}^l + \mathcal{L}_\text{contrastive}(\{\mathbf{Q}_{k}^l\}_{k=1}^K)$;
\ENDIF
\STATE $\mathcal{L} \gets \mathcal{L}+\mathcal{L}^l$;
\ENDFOR
\STATE \textbf{Backward pass:}
\STATE Update networks and $\mathbf{Q}$.
\end{algorithmic}
\label{algorithm1}
\end{algorithm}
\end{minipage}
\vspace{-35pt} 
\end{wrapfigure}
\section{Method}\label{method}

This section provides an overview of anti-overlapping object detection with DETR variants, focusing on the role of content queries and how contrastive learning refines their distributions. We then detail the proposed contrastive loss, including query partitioning and its inter- and intra-class components.

\subsection{Overview}
\cref{fig:fig3} illustrates our anti-overlapping detection framework using DETR variants. The backbone and encoder extract latent features from the input image, which are then queried by learnable content queries in the decoder to predict objects. To enhance anti-overlapping capability, MMCL clusters decoder queries into $K$ class-specific groups via a contrastive loss during training, where $K$ is the number of object categories. For multi-layer decoders, a subset of layers $T \subseteq L$ is selected for contrastive supervision, where $L$ is the decoder layer index set $\{\,l\in\mathbb{Z}\mid 0\le l\le 5\,\}$. At each target layer $l \in T$, content queries $\mathbf{Q}^l\in \mathbb{R}^{N\times 256}$ are evenly partitioned into $K$ groups $\{\mathbf{Q}_{k}^l\}_{k=1}^K$, where $\mathbf{Q}_{k}^l \in \mathbb{R}^{n\times 256}$ and $n=\lfloor N/K\rfloor$\footnote{In the case of an uneven split (i.e., $N\neq K \cdot n$), the first $r$ groups will be distributed $n+1$ queries, while remaining groups obtain $n$ queries, where $r=N-K\cdot n$.}. They are then refined with the contrastive loss $\mathcal{L}_\text{contrastive}$, while retaining guidance from the baseline loss $\mathcal{L}_{\text{base}}$. This ensures queries are discriminative across classes and effective at extracting features from overlapping objects, improving overall anti-overlapping detection performance. Algorithm~\ref{Algorithm_MMCL} summarizes the key steps in MMCL.

\subsection{Query Partition}

\begin{figure*}[t]
\centering
\includegraphics[width=\linewidth]{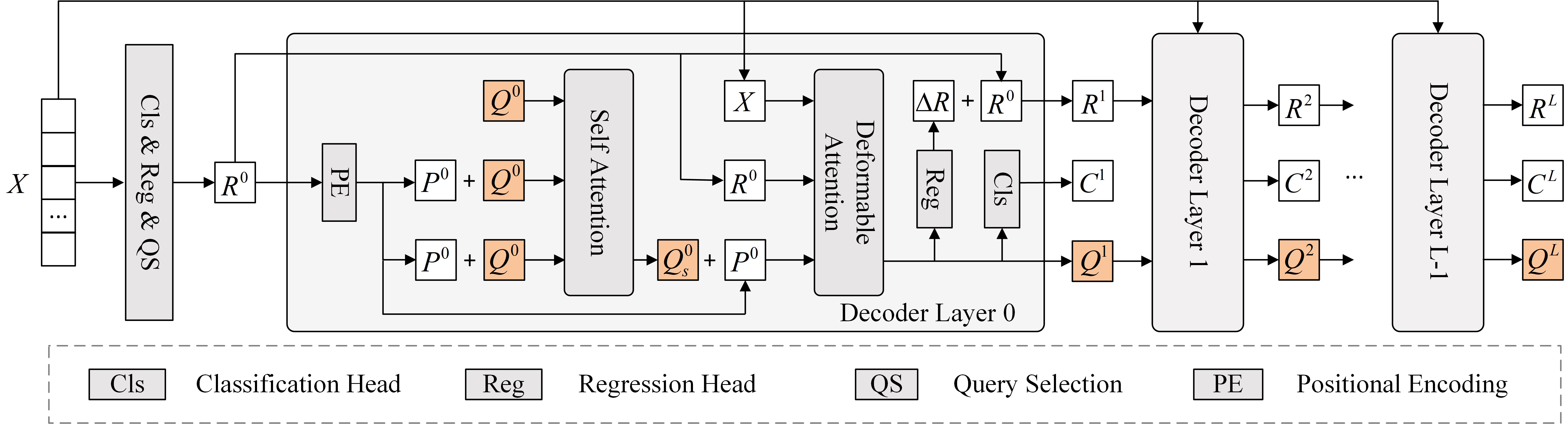}
\caption{Detailed illustration of the decoder’s content query mechanism in DINO~\cite{DINO}. After initializing the candidate boxes $R^0$ by the classification head, the regression head, and the query selection mechanism. Each decoder layer refines the content queries through self-attention and deformable attention, guided by classification head, regression head, and positional encoding mechanism~\cite{DINO}. The iterative update of content queries across layers enhances feature representation and detection accuracy.
Among the inputs of the decoder, \textbf{only} content queries $\mathbf{Q}^0$ are initiated independent of input feature $X$, which directly determine the final prediction results. Inspired by this, we propose MMCL to optimize their priors.
}
\label{fig:fig4}
\end{figure*}

In DETR-based detectors such as DINO~\cite{DINO} (\cref{fig:fig4}), content queries play a central role in decoding object information from image features. Given an input image, the backbone and encoder extract multi-scale spatial features $\mathbf{X}$, which are utilized as the decoder input and to predict initial reference boxes $\mathbf{R}^0$ as the prior of the initial positional queries $\mathbf{P}^0$~\cite{DINO}. Content queries $\mathbf{Q}^0$ are randomly initialized learnable embeddings independent of input image and positional queries.
The $l$-th decoder layer can be simplified as follows:
\begin{equation}
\label{equation decoder in appendix}
\mathbf{Q}^{l+1},\mathbf{R}^{l+1},\mathbf{C}^{l+1}=\mathcal{D}^l(\mathbf{Q}^l,\mathbf{R}^l,\mathbf{X};\theta^l),
\end{equation}
where $\mathcal{D}$ represents the decoder layer, $\mathbf{C}$ the classification results, and $\theta$ the learnable parameters.
Through multiple decoder layers, self-attention combines content and positional queries to extract global object features, while deformable attention uses reference boxes $\mathbf{R}^l$ to focus queries on relevant spatial regions. Linear mapping layers subsequently predict classification scores and bounding box refinements. 
This process demonstrates that content queries directly encode both class and localization information and determine the quality of the prediction result, whereas positional queries mainly provide spatial priors.




The direct influence of content queries on prediction results motivates us to structure their prior distribution more explicitly.
Instead of leaving the query priors unconstrained, we partition them into class-specific clusters so that each group is endowed with a natural class preference during prediction.
Under the intra-class attraction effects of the additional $L_\text{contrastive}$, intra-group content queries gradually tend to the same class preference, although queries may match ground-truth objects of distinct classes in the early phase. 
Meanwhile, the inter-class repulsion effects of $L_\text{contrastive}$ encourage inter-group content queries to evolve divergent class preferences.
As training progresses, $L_\text{base}$ stabilizes group-class and query-object associations, progressively specializing each group of queries into class-specific ones (i.e., intra-class queries), thereby increasing the effective number of training samples per class. Consequently, intra-class content queries learn to extract discriminative features for specific categories even in heavily overlapping scenarios. This category-aware clustering enhances inter-class separability while preserving intra-class diversity, ultimately improving anti-overlapping detection performance in complex X-ray images.

\subsection{Contrastive Loss}

\begin{figure}[t]
    \centering
    \includegraphics[width=0.6\linewidth]{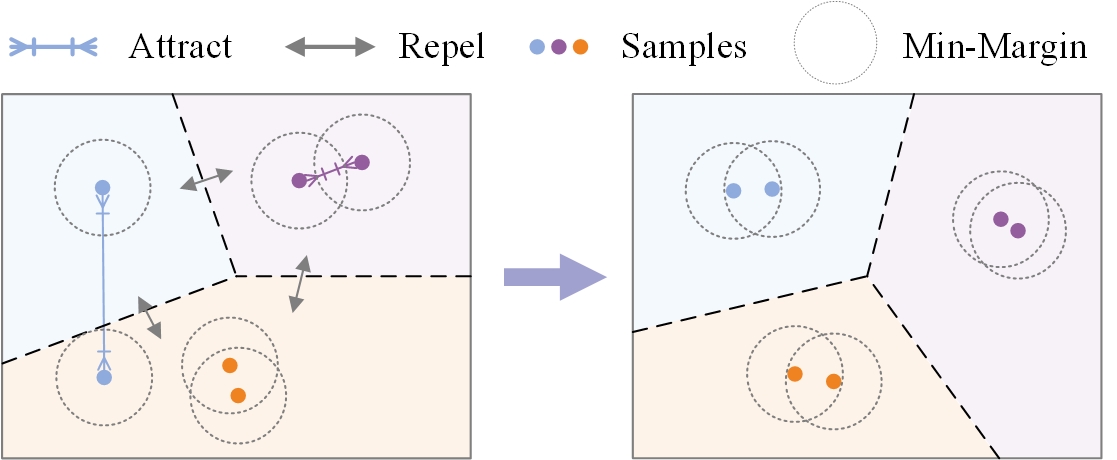}
    \caption{Illustration of how the proposed contrastive loss adjusts the distribution of content queries. The loss simultaneously repels inter-class samples and attracts intra-class samples, promoting clearer class separation. Each sample attracts only those intra-class samples lying outside a defined minimum-margin radius, thereby maintaining appropriate intra-class diversity. Points of the same color denote samples from the same class, while points of different colors represent content queries from different classes.}
    \label{fig:fig5}
\end{figure}

A typical contrastive loss consists of two complementary components: an intra-class term that maintains diversity within a category and an inter-class term that enforces separability between categories. The proposed contrastive loss is formulated as follows:
\begin{equation}
\label{equation MMCxx}
\small
\mathcal{L}_{\text{MMCL}}(\{\mathbf{Q}_{k}\}_{k=1}^K )= \gamma\mathcal{L}_{\text{IMC}}(\{\mathbf{Q}_{k}\}_{k=1}^K) + \eta \mathcal{L}_{\text{IME}}(\{\mathbf{Q}_{k}\}_{k=1}^K),
\end{equation}
where $\mathcal{L}_{\text{IMC}}$ and $\mathcal{L}_{\text{IME}}$ denote the Intra-class Min-margin Clustering (IMC) loss and the Inter-class Moderate Exclusion (IME) loss, respectively. The weighting coefficients $\gamma$ and $\eta$ control the trade-off between intra-class diversity and inter-class separability of content queries. \cref{fig:fig5} illustrates the changes in the distributions of content queries before and after applying the proposed contrastive loss.

\textbf{Inter-Class Term.} 
The inter-class moderate exclusion (IME) loss $\mathcal{L}_\text{{IME}}$ penalizes excessive similarity between samples from different categories by minimizing the mean cross-entropy of their cosine similarities. It is defined as:
\begin{equation}
\label{equation inter}
\begin{split}
\mathcal{L}_\text{{IME}}(\{\mathbf{Q}_{k}\}_{k=1}^K)=-\M{\mathds{1}{[k_1 \neq k_2]} \cdot \log \paren{1-s_{i,j}^{k_1,k_2}}},
\end{split}
\end{equation}
\begin{equation}
\label{equation inter 2}
s_{i,j}^{k_1,k_2}=\max \paren*{0, \frac{\mathbf{q}_i^{k_1}\cdot \mathbf{q}_j^{k_2}}{\norm{\mathbf{q}_i^{k_1}}_2\cdot \norm{\mathbf{q}_j^{k_2}}_2}},
\end{equation}
where $\mathds{1}{[k_1 \neq k_2]}$ is an indicator function that equals 1 when $k_1 \neq k_2$, and 0 otherwise. $\mathbb{M}$ denotes the empirical mean over all indexed query pairs. $s_{i,j}^{k_1,k_2}$ measures the similarity between the $i$-th query of class $k_1$ and the $j$-th query of class $k_2$. Notably, the cosine similarity term $s_{i,j}^{k_1,k_2}$ is truncated within the range $[0, 1]$, allowing the model to focus on challenging samples while reducing the risk of overfitting to easily distinguishable ones. Specifically, for an inter-class pair ($k_1\neq k_2$), as $s_{i,j}^{k_1,k_2}\!\to1$, the loss $\mathcal{L}_{\mathrm{IME}}\!\to+\infty$; since
$\partial\mathcal{L}_{\mathrm{IME}}/\partial s = 1/(1-s) > 0$,
gradient descent decreases $s$, thereby reducing inter-class similarity. In practice, backpropagation through the IME loss drives  $s_{i,j}^{k_1,k_2}$ toward its lower bound (approximately 0) until a balance is established with the other loss components.

\textbf{Intra-Class Term.} To attract samples within the same class while preserving sufficient distinctiveness and diversity, the proposed intra-class min-margin clustering (IMC) loss $\mathcal{L}_\text{{IMC}}$ introduces a minimum margin $m$, defined as:
\begin{align}
\label{equation intra 1}
\mathcal{L}_{\mathrm{IMC}}(\{\mathbf{Q}_{k}\}_{k=1}^K)=-\M{\mathcal{M}_{i,j}^k \bracket{w_{i,j}^k \cdot \log(s_{i,j}^{k,k})}} ,
\end{align}
\begin{align}
\label{equation intra 2}
\mathcal{M}_{i,j}^k=
\begin{cases}
1,  & w_{i,j}^k \cdot \log(s_{i,j}^{k,k})\ge m \\
0, & w_{i,j}^k \cdot \log(s_{i,j}^{k,k})< m,
\end{cases}
\end{align}
where $w_{i,j}^{k}=\exp(-\alpha \cdot \mathrm{rank}(s_{i,j}^{k,k}))$ weights the loss based on the similarity ranking among intra-class sample pairs, emphasizing similar pairs to accelerate training~\cite{C2AM}.
$\mathcal{M}\in \mathds{R}^{K\times n\times n}$ is a mask that sets elements to 0 when the weighted similarity loss of a sample, $w_{i,j}^k \cdot \log(s_{i,j}^{k,k})$, falls below the intra-class minimum margin $m$.
Consequently, the IMC loss only attracts sample pairs whose weighted similarity exceeds $m$. A larger $m$ allows greater diversity among intra-class samples.
Since $0\le w_{i,j}\le 1$, if all $\log(s_{i,j}^{k,k})< m$, the loss achieves the global optimum with $\mathcal{L}_\mathrm{IMC}=0$. Overall, the IMC loss mitigates excessive homogeneity and preserves intra-class diversity in content queries through the adjustable hyperparameter $m$.

\subsection{Discussion}

To analyze the advantages of the proposed contrastive loss for multi-class, multi-sample exclusion and clustering, we compare it with a classic loss (N-pair~\cite{N-pair}) and two recent losses (OCA~\cite{OCA_loss} and IIC~\cite{IIC_loss}), while other related losses are evaluated experimentally in~\Cref{tab:table9}.
The N-pair loss and OCA loss are formulated as:
\begin{equation}
\label{equation N-pair}
\mathcal{L}_{\text{N-pair}}=-\mathbb{M}\left[\log\left(\frac{ e^{s_{i,j}^{k,k}}}{ e^{s_{i,j}^{k,k}}+e^{s_{i,j}^{k_1,k_2}}}\right)\right],
\end{equation}
\begin{equation}
\label{equation OCA}
\mathcal{L}_{\text{OCA}}=-\mathbb{M}\left[\log\left(\frac{ e^{s_{i,j}^{k,k}}}{ e^{s_{i,j}^{k,k}}+e^{s_{i,j}^{k_1,k_2}}+\mathds{1}_{[s<\tau]}e^{s_{i,j}^{k,k}}}\right)\right],
\end{equation}
where $k_1 \neq k_2$. N-pair loss simultaneously attracts intra-class samples $s_{i,j}^{k,k}$ and repels inter-class samples $s_{i,j}^{k_1,k_2}$, which can lead to imbalance during training. OCA loss introduces a term $\mathds{1}{[s<\tau]} e^{s{i,j}^{k,k}}$ to orthogonalize intra-class pairs below a threshold, but it does not prevent intra-class homogeneity. IIC loss~\cite{IIC_loss} is defined as:
\begin{equation}
\label{equation IIC}
\mathcal{L}_{\mathrm{IIC}}=\mathbb{M}\Big[\Big( \KL{q^{k}_i}{q^{k}_j} + \KL{q^{k}_j}{q^{k}_i}\Big)- \paren*{ \KL{q^{k_1}_i}{q^{k_2}_j} + \KL{q^{k_2}_j}{q^{k_1}_i}}\Big],
\end{equation}
where $D_{\mathrm{KL}}(\cdot)$ is the Kullback-Leibler divergence. Without a min-margin mechanism, intra-class queries can collapse, leading to homogeneity.

In contrast, the proposed contrastive loss decouples attraction and repulsion using IMC and IME losses.\textit{ Its advantages include: (i) flexible weighting between the two components balances the intra-class attraction and inter-class repulusion tasks, and (ii) the adjustable margin $m$ in IMC preserves suitable intra-class diversity.} These properties make the proposed contrastive loss particularly suitable for multi-class, multi-sample exclusion and clustering, and essential for improving anti-overlapping detection in X-ray images. Experimental comparisons with other contrastive losses of similar style, including N-pair loss~\cite{N-pair}, InfoNCE loss~\cite{InfoNCE}, OCA loss~\cite{OCA_loss}, IIC loss~\cite{IIC_loss}, and ICE loss~\cite{ICE_loss}, are provided in~\Cref{tab:table9}.

\section{Experiments}
\label{Experiments}

In this section, we first describe the datasets, evaluation metrics, and implementation details of the proposed method. Subsequently, we assess the generalization and effectiveness of MMCL across four DETR variants and two backbone networks on PIXray~\cite{PIXray} and OPIXray~\cite{OPIXray} datasets. We then compare the proposed contrastive loss with five widely used contrastive losses within the MMCL framework. Extensive ablation studies on PIXray~\cite{PIXray} further investigate the selection of target decoder layers and hyperparameters of the proposed loss. Additionally, we provide visualizations of prediction results and sampling points, alongside an analysis of model complexity, to offer a comprehensive understanding of the proposed method for anti-overlapping object detection.
Finally, we analyze the effect of MMCL on anti-overlapping ability on the PIDray~\cite{PIDray} dataset, which includes three subsets with different overlapping degrees.

\subsection{Datasets and Metrics}
We conduct experiments on three widely used X-ray object detection datasets.
\textbf{PIXray}~\cite{PIXray}: This dataset includes 5,046 X-ray images with 15 classes prohibited items annotated as instance-level masks. For our experiments, we convert the annotations to COCO-style~\cite{COCO} bounding box labels for the prohibited item detection task.
\textbf{OPIXray}~\cite{OPIXray}: This dataset contains 8,885 X-ray images, with 7,019 for training and 1,776 for testing, covering five categories of cutters: folding knife (FO), straight knife (ST), scissor (SC), utility knife (UT), and multi-tool knife (MU). 
\textbf{PIDray}~\cite{PIDray}: This dataset comprises 47,677 X-ray images spanning 12 categories of prohibited items, with 29,457 images for training and 18,220 for testing. To analyze the impact of object overlapping degree, the test set is further divided into three difficulty levels: Easy (9,482 images, single item), Hard (3,733 images, multiple items), and Hidden (5,005 images, intentionally concealed items).

For the PIXray and PIDray datasets, we adopt the COCO~\cite{COCO} evaluation metric, including AP$_{50}$, AP$_{75}$, and mean average precision (AP). AP measures the detector’s precision across multiple IoU thresholds, reflecting overall detection performance. AP$_{50}$ and AP$_{75}$ evaluate precision at IoU thresholds of 0.5 and 0.75, respectively, representing moderate and strict localization accuracy. mAP is computed as the average AP across all categories, providing a comprehensive assessment of both accuracy and recall. For the OPIXray~\cite{OPIXray} dataset, we adopt the VOC~\cite{VOC} evaluation metric. AP is computed per category at an IoU threshold of 0.5, and mAP is obtained by averaging across all categories, providing a holistic measure of detector performance in both localization and classification.

\subsection{Implementation Details}
All training and testing are performed on the same platform equipped with an NVIDIA GeForce RTX 4090 GPU, an Intel Core i9-13900K CPU, 64 GB of memory, Windows 10, and PyTorch 1.13.1. To control for non-parameter factors, we employ pre-trained models provided on the official MMDetection website, including ResNet-50 and Swin-L backbones. Transformer-based models such as DINO are trained with the AdamW optimizer, using a learning rate of 1e-4 and a weight decay of 1e-4. All models are trained for 12 epochs by default, following their original training protocols, with an input image size of $320\times320$.

\begin{table*}[!ht]
\caption{Comparison with state-of-the-art general detectors on PIXray~\cite{PIXray}. ``-" indicates that the corresponding data are not disclosed or cannot be obtained because the models are not publicly available. \#Q represents the number of object queries.}
\footnotesize
\begin{tabularx}{\linewidth}{l|cAccAc|AAAAA}
\toprule
Method & Backbone & FPS & \#Params & GFLOPs & \#Q & AP & AP$_{50}$ & AP$_{75}$ & AP$_\text{S}$ & AP$_\text{M}$ & AP$_\text{L}$\\
\midrule
Faster R-CNN~\cite{Faster}& ResNeXt-101 & 70& 59.83M & 28.35&- & 53.6 & 82.3 & 60.8 & 3.9 & 37.7 & 62.7 \\
Mask R-CNN~\cite{Mask}& ResNeXt-101 & 73& 60.04M & 28.35&-& 52.4 & 81.9 & 59.4 & 4.2 & 36.2 & 61.3 \\
Cascade R-CNN~\cite{Cascade}&ResNet-50 & 61 & 68.97M &22.37&-&56.5&81.3&63.2&8.0&41.0&65.9\\
ATSS~\cite{ATSS}&ResNet-101 &66&51.14M&27.82&-&52.8&80.8&60.2&7.0&37.4&63.6\\
GFLv1~\cite{GFLv1}&ResNeXt-101 &66&50.70M&28.51&-&57.5&82.8&66.0&9.1&42.0&67.4\\
\midrule
DETR~\cite{Deformable-DETR}& ResNet-50 & 60 & 52.14M & 13.47 & 300 & 44.6 & 74.2 & 48.5 & 9.6 & 30.0 & 53.0\\
\rowcolor{cyan} DETR~\cite{Deformable-DETR} + MMCL & ResNet-50 & 60 & 52.14M & 13.47 & 300 & 48.4 \bl{(+3.8)} & 76.9 & 52.3 & 9.1 & 34.3 & 57.5\\
\midrule
RT-DETR~\cite{RT-DETR}& ResNet-50 & 64 & 42.81M & 17.07 & 60 & 62.3 & 85.3 & 69.9 & 25.6 & 48.0 &70.9\\
\rowcolor{cyan} RT-DETR~\cite{RT-DETR} + MMCL & ResNet-50 & 64 & 42.81M & 17.07 & 60 & 63.6 \bl{(+1.3)} & 85.9 & 71.4 & 24.0 & 49.9 &72.6\\
\midrule
DINO~\cite{DINO}& ResNet-50 & 54 & 58.38M & 26.89 & 30 & 64.3 & 86.5 & 71.0 & 19.3 & 48.9 & 73.9\\
\rowcolor{cyan} DINO~\cite{DINO} + MMCL& ResNet-50 & 54 & 58.38M & 26.89 & 30 & 66.7 \bl{(+2.4)} & 87.5 & 74.4 & 23.5 & 50.7 &75.5\\
DINO~\cite{DINO} & Swin-L & 40 & 229.0M & 156.0 & 30 & 72.8 & 90.0 & 80.1 & 38.3 & 60.4 & 80.4\\
\rowcolor{cyan} DINO~\cite{DINO} + MMCL& Swin-L & 40 & 229.0M & 156.0 & 30 & 73.2 \bl{(+0.4)} & 89.7 & 79.9 & 37.4 & 58.9 &81.1\\    \midrule
AO-DETR~\cite{AO-DETR}& ResNet-50 & 54 & 58.38M & 26.89 & 30 & 65.6 & 86.1 & 72.0 & 23.9 & 50.7 & 74.8\\
\rowcolor{cyan} AO-DETR~\cite{AO-DETR} + MMCL& ResNet-50 & 54 & 58.38M & 26.89 & 30 & 66.8 \bl{(+1.2)} & 87.6 & 74.3 & 24.1 & 52.4 &75.9\\
AO-DETR~\cite{AO-DETR}& Swin-L & 40 & 229.0M & 156.0 & 30 & 73.9 & 89.9 & 80.6 & 40.5 & 62.4 & 81.6\\
\rowcolor{cyan} AO-DETR~\cite{AO-DETR} + MMCL& Swin-L & 40 & 229.0M & 156.0 & 30 & 74.6 \bl{(+0.7)} &90.6 &81.6 &39.3 &62.6 &82.2\\
\bottomrule
\end{tabularx}
\label{tab:table1}
\end{table*}
\begin{table*}[ht]
\caption{Comparison with state-of-the-art general detectors on OPIXray~\cite{OPIXray}. ``-" indicates that the corresponding data are not disclosed or cannot be obtained because the models are not publicly available. \#Q represents the number of object queries.}
\centering
\footnotesize
\begin{tabularx}{\linewidth}{l|cAccAc|AAAAA}
\toprule
Method & Backbone & FPS & \#Params & \#GFLOPs & \#Q & mAP&FO& ST& SC& UT& MU\\
\midrule
Faster R-CNN~\cite{Faster}& ResNeXt-101 & 70& 59.83M & 28.35&- & 73.4&80.6 & 45.4 & 89.1  & 69.1 & 83.1 \\
Mask R-CNN~\cite{Mask}& ResNeXt-101 & 73& 60.04M & 28.35& - & 77.2 & 83.6 & 55.9 & 89.8 & 71.5 & 85.2 \\
Cascade R-CNN~\cite{Cascade}&ResNet-50 &61&68.97M&22.37&-&72.8&75.7&50.0&89.4&70.0&79.0\\
ATSS~\cite{ATSS}&ResNet-101 &66&51.14M&27.82&-&67.5&72.8&38.0&88.6&58.0&80.2\\
GFLv1~\cite{GFLv1}&ResNeXt-101 &66&50.70M&28.51&-&75.6&80.0&53.6&89.3&71.7&83.4\\
\midrule
DETR~\cite{Deformable-DETR}& ResNet-50 & 60 & 52.14M & 13.47 & 20 & 52.4 & 51.2 & 21.5 & 81.6 & 49.0&58.4\\
\rowcolor{cyan} DETR~\cite{Deformable-DETR} + MMCL & ResNet-50 & 60 & 52.14M & 13.47 & 20 & 58.5 \bl{(+6.1)} & 61.5 & 23.8 & 85.4 & 47.8 & 74.8\\
\midrule
RT-DETR~\cite{RT-DETR}& ResNet-50 & 64 & 42.81M & 17.07 & 320 & 61.8 & 61.1 & 26.0 & 88.6 & 56.4 &76.8\\
\rowcolor{cyan} RT-DETR~\cite{RT-DETR} + MMCL& ResNet-50  & 64 & 42.81M & 17.07 & 320 & 62.5 \bl{(+0.7)} & 65.9 & 22.3 & 86.4 & 57.1 &80.7\\
\midrule
DINO~\cite{DINO}& ResNet-50 & 54 & 58.38M & 26.89 & 30 & 78.2 & 83.2 & 58.8 & 89.4 & 72.7 & 86.7\\
\rowcolor{cyan} DINO~\cite{DINO} + MMCL & ResNet-50 & 54 & 58.38M & 26.89 & 30 & 78.6 \bl{(+0.4)} & 83.9 & 57.2 & 90.4 & 74.2 &87.1\\
DINO~\cite{DINO}& Swin-L & 40 & 229.0M & 156.0 & 30 & 80.0& 84.2& 61.1& 89.0 &78.9& 86.6\\
\rowcolor{cyan} DINO~\cite{DINO} + MMCL & Swin-L & 40 & 229.0M & 156.0 & 30 & 81.8 \bl{(+1.8)} & 86.9 & 64.7 & 89.8 & 78.9 &88.9\\    \midrule
AO-DETR~\cite{AO-DETR}& ResNet-50 & 54 & 58.38M & 26.89 & 30 & 79.2 & 83.8 & 60.5 & 90.1 & 74.7 & 87.1\\
\rowcolor{cyan} AO-DETR~\cite{AO-DETR} + MMCL & ResNet-50 & 54 & 58.38M & 26.89 & 30 & 80.3 \bl{(+1.1)} & 84.6 & 63.6 & 90.2 & 74.9 &88.0\\
AO-DETR~\cite{AO-DETR}& Swin-L & 40 & 229.0M & 156.0 & 30 & 80.8 & 84.8 & 63.0 & 90.1 & 77.7 & 88.4\\
\rowcolor{cyan} AO-DETR~\cite{AO-DETR} + MMCL & Swin-L & 40 & 229.0M & 156.0 & 30 & 82.1 \bl{(+1.3)} &87.4 &63.9 &89.9&79.3 &89.8\\
\bottomrule
\end{tabularx}
\label{tab:table2}
\end{table*}
 
\subsection{Main Results}
\textbf{General Object Detectors.} 
The results in Tables~\ref{tab:table1} and~\ref{tab:table2} collectively demonstrate the consistent effectiveness and strong generalization ability of the proposed MMCL across different detection frameworks and backbone networks. When incorporated into DETR~\cite{DETR}, RT-DETR~\cite{RT-DETR}, DINO~\cite{DINO}, and AO-DETR~\cite{AO-DETR}, MMCL leads to consistent performance gains, improving AP by up to +3.8\% on PIXray~\cite{PIXray} and mAP by up to +6.1\% on OPIXray~\cite{OPIXray}. These enhancements confirm that MMCL effectively mitigates feature confusion arising from overlapping objects in complex X-ray imagery. Furthermore, the improvement trends remain stable across various object scales and backbone architectures, from ResNet-50~\cite{ResNet} to Swin-L~\cite{Swin-Transformer}, underscoring MMCL's adaptability to both convolutional and transformer-based detectors. Notably, AO-DETR + MMCL achieves the highest performance on both datasets (74.6\% AP on PIXray and 82.1\% mAP on OPIXray), surpassing recent state-of-the-art models. Overall, MMCL serves as a versatile enhancement mechanism that substantially improves feature representation and robustness in X-ray contraband detection.

\textbf{Prohibited Item Detectors.} 
The results in Table~\ref{tab:table3} demonstrate that integrating MMCL into AO-DETR substantially enhances performance compared with existing prohibited item detectors on the OPIXray dataset. AO-DETR (Swin-L) with MMCL achieves the highest mean average precision (mAP) of 89.8\%, surpassing all prior methods, including specialized detectors such as LAreg~\cite{CLCXray}, LAcls~\cite{CLCXray}, and POD-F-X~\cite{POD}. These improvements are consistent across all item categories (FO, ST, SC, UT, and MU), confirming the robustness and adaptability of MMCL under complex X-ray inspection conditions. Moreover, the MMCL-equipped AO-DETR (ResNet-50) achieves a notable mAP of 87.6\%, surpassing other detectors with comparable backbones, such as DOAM~\cite{OPIXray} and XDet~\cite{Xdet}. The consistent performance gains across both lightweight and transformer-based variants highlight the generality of MMCL in mitigating feature entanglement from overlapping objects, leading to more accurate and stable detection of concealed contraband.

\begin{table*}[!ht]
\caption{Comparison with state-of-the-art prohibited item detectors on OPIXray~\cite{OPIXray}. ``-" indicates that the corresponding data are not disclosed or cannot be obtained because the models are not publicly available.}
\footnotesize
\begin{tabularx}{\linewidth}{l|cAc|A|AAAAA|rA}
\toprule
Method & Backbone&Epoch&Input Size &mAP &FO& ST& SC& UT& MU&\#Params&FPS\\
\hline
DOAM~\cite{OPIXray}&ResNet-50&-&-&82.4& 86.7 &68.6 &90.2& 78.8& 87.7&90.79 M& -\\
DOAMv2~\cite{OPIXrayv2}&ResNet-50&-&-& 83.8 &87.6 &72.7& 90.0& 80.8 &87.8&90.79 M&-\\
FCOS + LIM~\cite{tao2021towards}&ResNet-50&-&-& 83.1&86.6&71.9&90.3&79.9&86.8 &-&-\\
XDet~\cite{Xdet}&ResNet-50&-&1280&86.7 &90.4 &76.0 &91.5&84.3 &91.3&41.19 M&25\\
LAreg~\cite{CLCXray}& ResNet-50&12&1280&87.4&\textbf{92.8}&71.2&96.6&83.5&92.9&-&-\\
LAcls~\cite{CLCXray}& ResNet-50&12&1280&88.3&90.0&75.0&97.6&85.7&\textbf{92.9}&-&-\\
DML-Net~\cite{yang2024dual}&ResNet-50&120&-&52.7&38.1&42.7&54.7&38.9&35.9&36.20 M&56\\
POD-F-R~\cite{POD}&ResNet-50&24&1333&84.9&88.7&76.0&88.9&82.8&88.1&118.32 M&7\\
POD-F-X~\cite{POD}&ResNeXt-50&24&1333&86.1&89.4&78.7&90.6&83.3&88.7&119.67 M&6\\
GADet-S~\cite{GADet}& CSP v5 &60&320&69.6& 72.6& 43.6&86.6& 67.5& 77.5&8.94 M&116\\
GADet-L~\cite{GADet}& CSP v5 &60&320&77.7& 81.8& 54.0&89.8& 77.5& 85.2&54.16 M&75\\
GADet-X~\cite{GADet}& CSP v5 &60&320&78.1& 83.1& 56.3&89.8& 75.7& 85.5&99.01 M&56\\
FDTNet~\cite{FDTNet}& ResNeXt-101&12 &512&82.0&87.9&60.2&96.1&78.9&87.1&66.17 M&-\\
FDTNet~\cite{FDTNet}& ResNeXt-101&12 &1333&88.0&91.5&74.6&97.6&85.2&91.2&66.17 M&-\\
Mix-Paste + LLS~\cite{chen2024augmentation}&ResNet-50&24&-& 83.7&-&-&-&-&-&-&19\\
AO-DETR~\cite{AO-DETR} & ResNet-50&15&640&87.2& 90.0& 80.1& 90.8& 85.6 &89.5&58.38 M&29\\
AO-DETR~\cite{AO-DETR} & Swin-L&15&640&89.0& 89.4& 80.4 &97.8& \textbf{87.4}& 90.0&229.0 M&15\\
\hline
\rowcolor{cyan} AO-DETR~\cite{AO-DETR} + MMCL &ResNet-50&15&640&87.6 &89.4& 82.7 &90.9& 85.0& 89.8&58.38 M&29\\
\rowcolor{cyan} AO-DETR~\cite{AO-DETR} + MMCL & Swin-L&15&640&\textbf{89.8}& 89.2& \textbf{84.0}& \textbf{99.4} &86.2 &90.1&229.0 M&15\\
\bottomrule
\end{tabularx}
\label{tab:table3}
\end{table*}

\textbf{Discussion.} To evaluate the efficacy and generalizability of the proposed MMCL mechanism across different DETR variants, four DETR-like models—DETR~\cite{Deformable-DETR}, RT-DETR~\cite{RT-DETR}, AO-DETR~\cite{AO-DETR}, and DINO~\cite{DINO}—are selected as baselines, together with two distinct backbone architectures, ResNet-50~\cite{R-CNN} and Swin-L~\cite{Swin-Transformer}. As shown in~\Cref{tab:table1}, incorporating MMCL into ImageNet-pretrained ResNet-50 models significantly enhances the AP of DETR, RT-DETR, DINO, and AO-DETR by 3.8\%, 1.3\%, 2.4\%, and 1.2\%, respectively, on the PIXray dataset, highlighting the robust generalization of the proposed mechanism. Similarly, with the ImageNet-pretrained Swin-L backbone, MMCL further improves the AP of DINO and AO-DETR by 1.8\% and 1.3\%, respectively, on the OPIXray dataset (as shown in~\Cref{tab:table2}), demonstrating its adaptability across architectures. Overall, MMCL consistently improves detection precision on both PIXray and OPIXray without increasing GFLOPs or parameter counts, while maintaining inference speed, confirming its efficiency and broad applicability.

\subsection{Ablation Studies}\label{section_ablation}
In this section, a series of ablation studies are conducted on the PIXray dataset to systematically evaluate the design and effectiveness of the proposed MMCL mechanism. First, we examine the influence of the inserted layer set $T$ on detection performance to determine the optimal configuration. Next, we analyze the compatibility and individual contributions of the inter-modality enhancement (IME) and intra-class min-margin clustering (IMC)  components. Finally, we investigate the impact of key hyperparameters on overall performance, including the minimum margin $m$ in the IMC loss and the weighting factors $\gamma$ and $\eta$ for the IMC and IME losses, respectively. All experiments in this section are performed using the vanilla DINO model with a ResNet-50 backbone.

\textbf{Ablation Study on Target Layer.}
\Cref{tab:table4} presents an ablation study evaluating the effect of different target layer settings $T$ on model performance, with all other hyperparameters fixed ($m=0.1$, $\eta=1$, and $\gamma=1$). The baseline model without target-layer supervision achieves an AP of 64.3\%. Introducing supervision at the first decoder layer ($T={0}$) yields the best results, with AP of 65.7\%, AP$_{50}$ of 87.1\%, and AP$_{75}$ of 72.7\%. This suggests that guiding the model at the earliest decoding stage facilitates more effective feature learning and enhances detection accuracy. Conversely, applying supervision to deeper layers ($ T={1}$ or $T={5}$) leads to diminished performance, particularly at $T={5}$, where optimization becomes unstable. Simultaneously supervising all decoder layers ($T=L$) produces moderate improvements but does not surpass the single-layer case. These findings highlight the importance of early-layer supervision under consistent training conditions.

\begin{table}[!ht]
\begin{center}
\footnotesize
    \caption{Ablation study on target layer.}
    \label{tab:table4}
    \begin{tabularx}{0.6\linewidth}{A|AAA}
    \toprule
     $T$  & AP & AP$_{50}$ & AP$_{75}$\\
    \midrule
    - &64.3 & 86.5 & 71.0\\
     \rowcolor{cyan} \{0\}&\textbf{65.7} &\textbf{87.1} &\textbf{72.7}\\
     \{1\}&64.5 &85.7 &71.5\\
     \{5\}&10.2&19.4 &9.5\\
     $L$ &64.9&86.3 &72.4\\
    \bottomrule
  \end{tabularx}
\end{center}
\end{table}

\textbf{Ablation Study on IMC and IME Losses.} 
\Cref{tab:table5} presents an ablation study assessing the impact of the IME ( for inter-class separability) and IMC (for intra-class diversity) losses on model performance. The baseline model without either component achieves an AP of 64.3\%. Introducing the IME loss alone improves the AP to 65.3\% and yields the highest AP$_{75}$ (72.9\%), indicating that enhancing inter-class separability helps the model better distinguish between object categories. The IMC loss alone also improves performance (AP=65.1\%), suggesting that encouraging intra-class diversity benefits feature robustness. When both losses are jointly applied, the model achieves the best overall performance (AP=65.7\%, AP$_{50}$=87.1\%, AP$_{75}$=72.7\%). These findings confirm that IME and IMC are complementary objectives that jointly enhance the model’s discriminative capability and generalization.

\begin{table}[!ht]
\footnotesize
\begin{center}
    \caption{Ablation study on IMC and IME losses.}
    \label{tab:table5}
    \begin{tabularx}{0.75\linewidth}{AA|AAA}
    \toprule
     IME & IMC & AP & AP$_{50}$ & AP$_{75}$\\
    \midrule
     \ding{55}&\ding{55}&64.3 & 86.5 & 71.0\\
    \ding{51}&\ding{55}&65.3 &86.7&\textbf{72.9}\\
    \ding{55}&\ding{51}&65.1 &86.7&72.1\\
    \rowcolor{cyan} \ding{51}&\ding{51}&\textbf{65.7} &\textbf{87.1} &72.7\\
    \bottomrule
  \end{tabularx}
\end{center}
\end{table}

\textbf{Ablation Study on Hyperparameters.}
We first set the default hyperparameters to $\eta = 1.0$ and $\gamma = 1.0$, and varied the margin $m$ to determine the optimal value $m^*$. As shown in~\Cref{Ablation experiments all} (left), the model achieves its highest performance when $m = 0.01$, reaching maximum AP, AP$_{50}$, and AP$_{75}$ scores of 66.4\%, 87.9\%, and 73.3\%, respectively. Fixing $\gamma = 1.0$ and using $m^* = 0.01$, we next searched for the optimal $\eta$. As presented in~\Cref{Ablation experiments all} (middle), the best results are obtained when $\eta = 0.5$, with AP, AP$_{50}$, and AP$_{75}$ values of 66.7\%, 87.5\%, and 74.4\%. Finally, with $m^* = 0.01$ and $\eta^* = 0.5$, we varied $\gamma$ and observed from~\Cref{Ablation experiments all} (right) that $\gamma = 1.0$ yields the highest AP (66.7\%) and AP$_{75}$ (74.4\%), outperforming $\gamma = 2.0$. In summary, the optimal hyperparameters for MMCL are $m^* = 0.01$, $\eta^* = 0.5$, and $\gamma^* = 1.0$.

\begin{table*}[!t]
  \centering
  \caption{Ablation study on $m$, $\eta$, and $\gamma$ on PIXray~\cite{PIXray}. 
  The superscript '$*$' denotes the optimal hyperparameter.}
  \footnotesize

  \subfloat[$\gamma=1$, $\eta=1$\label{tab:ablate_m}]{
      \centering
        \begin{tabularx}{0.35\linewidth}{c|cAA}
          \toprule
          $m$  & AP & AP$_{50}$ & AP$_{75}$ \\
          \midrule
          $1\times 10^{-1}$ &65.7 \bl{(+ 1.4)} &87.1 &72.7 \\
          $3\times 10^{-2}$ &66.2 \bl{(+ 1.9)}&87.2 &73.2  \\ 
          \rowcolor{cyan} $\mathbf{1\times 10^{-2}}$ &\textbf{66.4} \textbf{\bl{(+ 2.1)}} &\textbf{87.9} &\textbf{73.3} \\ 
          $3\times 10^{-3}$ &65.9 \bl{(+ 1.6)}&87.4 &73.2 \\
          $1\times 10^{-4}$ &60.0 \bl{(+ 1.7)}&82.2 &66.4 \\
          \bottomrule
        \end{tabularx}
  }
\hfill 
  \subfloat[$\gamma=1$, $m^*=0.01$\label{tab:ablate_eta}]{
      \centering
      \begin{tabularx}{0.3\linewidth}{c|cAA}
        \toprule
        $\eta$  & AP & AP$_{50}$ & AP$_{75}$ \\
        \midrule
        2.00   &66.0 \bl{(+ 1.7)}&87.2 &73.1 \\
        1.00   &65.7 \bl{(+ 1.4)}&86.7 &72.4\\
        \rowcolor{cyan} \textbf{0.50} &\textbf{66.7} \textbf{\bl{(+ 2.4)}}&\textbf{87.5} &\textbf{74.4} \\
        0.25&65.3 \bl{(+ 1.0)}&86.9 &73.2  \\
        0.10 &56.2 \bl{(+ 1.9)}&79.1 &63.1 \\
        \bottomrule
      \end{tabularx}
  }\hfill 
  \subfloat[$\eta^*=0.5$, $m^*=0.01$\label{tab:ablate_gamma}]{
      \centering
      \begin{tabularx}{0.3\linewidth}{c|ccc}
        \toprule
        $\gamma$ & AP & AP$_{50}$ & AP$_{75}$ \\
        \midrule
        5.0   & 65.1 \bl{(+ 0.8)}&86.8 &72.3 \\
        2.0 &66.6 \bl{(+ 2.3)}&\textbf{88.1} &73.7 \\
        \rowcolor{cyan} $\mathbf{1.0}$ &\textbf{66.7} \textbf{\bl{(+ 2.4)}}& 87.5 &\textbf{74.4} \\
        0.5 &66.5 \bl{(+ 2.2)}&87.6 &74.1  \\
        0.2 &66.1 \bl{(+ 1.8)}&87.5 &73.6 \\
        \bottomrule
      \end{tabularx}
  }

\label{Ablation experiments all}
\end{table*}

\begin{table}[!ht]
\footnotesize
\caption{Comparison of contrastive losses using the MMCL framework.}
\label{tab:table9}
\centering
\begin{tabularx}{1\linewidth}{l|AAAAAA}
\toprule
&  AP & AP$_{50}$ & AP$_{75}$ & AP$_\text{S}$ & AP$_\text{M}$ & AP$_\text{L}$ \\
\midrule
DINO & 64.3 & 86.5 &71.0 &19.3 &48.9 &73.9\\
DINO + IIC~\cite{IIC_loss} &64.6 &86.0 &71.3 &20.3 &49.7 &73.7\\
DINO + ICE~\cite{ICE_loss}&64.8 &86.4 &71.8 &19.8 &50.1 &73.9\\
DINO + N-pair~\cite{N-pair} & 65.4 &86.6 &72.7 &19.9 &50.2 &75.3\\
DINO + InfoNCE~\cite{InfoNCE} & 66.1 &87.4 &74.2 &19.4 &51.6 &75.5 \\
DINO + OCA~\cite{OCA_loss}&66.3 &86.8 &72.9 &18.4 &50.7 &\textbf{76.1}\\
\rowcolor{cyan} DINO + MMCL & \textbf{66.7} &\textbf{87.5} &\textbf{74.4} &\textbf{23.5} &\textbf{50.7} &75.5 \\
\bottomrule
\end{tabularx}
\end{table}

\textbf{Ablation Study on Generalization}. 
\Cref{tab:table9} compares MMCL with other contrastive loss variants in the DINO detector on the PIXray dataset~\cite{PIXray}. MMCL achieves the best overall performance, with an AP of 66.7\%, AP$_{50}$ of 87.5\%, and AP$_{75}$ of 74.4\%, while significantly improving small-object detection (AP$_\text{S}$ = 23.5\%). These improvements highlight the effectiveness of MMCL in addressing the overlapping-object problem inherent in X-ray imagery. By introducing a contrastive mechanism that explicitly corrects the content query distribution, MMCL enhances inter-class separability and preserves intra-class diversity, leading to more discriminative and balanced feature representations. Compared with other contrastive loss variants such as InfoNCE and OCA, MMCL achieves consistent gains without compromising large-object detection, demonstrating its robustness and adaptability for complex anti-overlapping X-ray detection tasks.

\begin{figure*}[!h]
    \centering
    \includegraphics[width=1\linewidth]{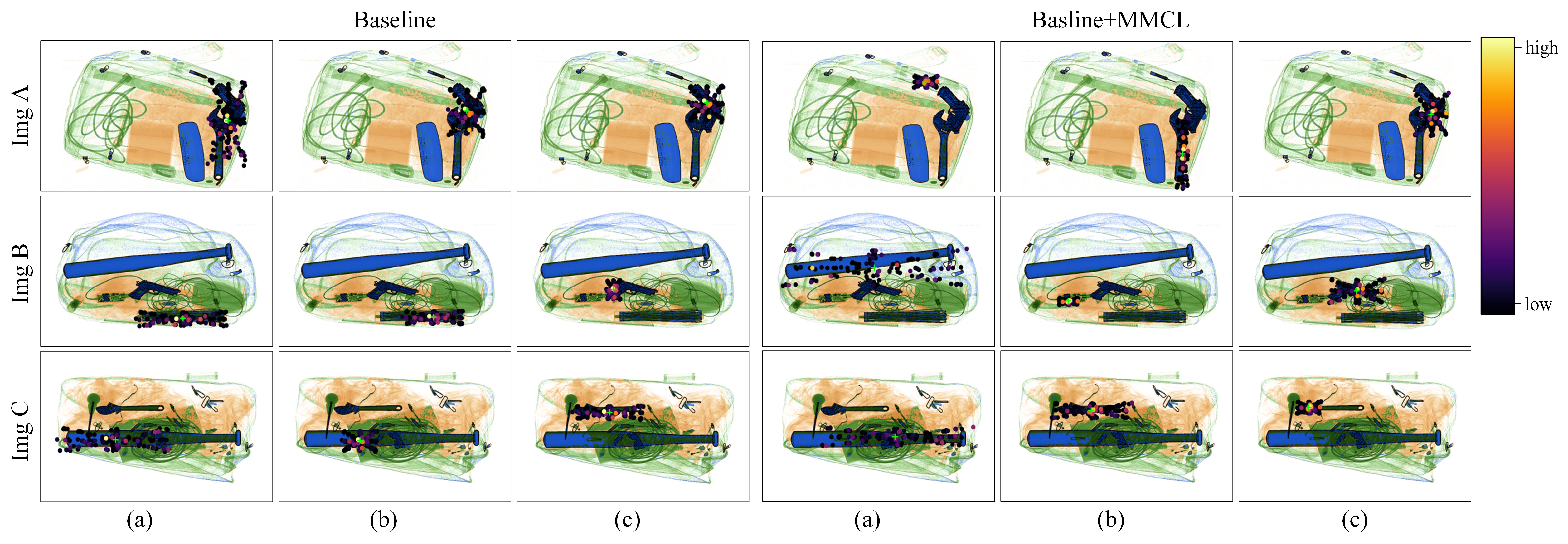}
    \caption{Visualization of representative sampling points. Columns (a), (b), and (c) correspond to the sampling points of the 4th, 23rd, and 25th queries, respectively. In the baseline model, these queries tend to attend to various background regions or unrelated objects. After integrating MMCL, the queries consistently focus on relevant prohibited items, such as the bat, wrench, and gun, respectively.}
    \label{fig:fig6}
\end{figure*}

\subsection{Visualization of Sampling Points}
\cref{fig:fig6} shows the sampling points~\cite{Deformable-DETR} in the $4$-th decoder layer of the DINO model, with and without MMCL integration. The PIXray dataset contains 15 categories, and we set the number of queries to 30. We use the bat, wrench, and gun—three prohibited items with distinct shapes—as examples. After MMCL integration, their content query group indices are the $2$-nd, $11$-th, and $12$-th. For clarity, we visualize the $4$-th, $23$-rd, and $25$-th queries from each group to analyze changes in category attributes before and after MMCL. In column (a), the $4$-th query of the baseline model lacks a fixed-category attribute, focusing on backgrounds in Img A and Img B and partially detecting the bat in Img C. With MMCL, the $4$-th query is assigned the bat category, consistently detecting it across images; in Img A, which contains no bat, it correctly focuses on the background. Columns (b) and (c) show similar behavior. Overall, the baseline queries have unstable category attributes. MMCL significantly increases the likelihood that a query consistently detects a specific category, thereby enhancing detection performance and stabilizing query category assignments.

\begin{figure*}[h]
    \centering
    \includegraphics[width=1\linewidth]{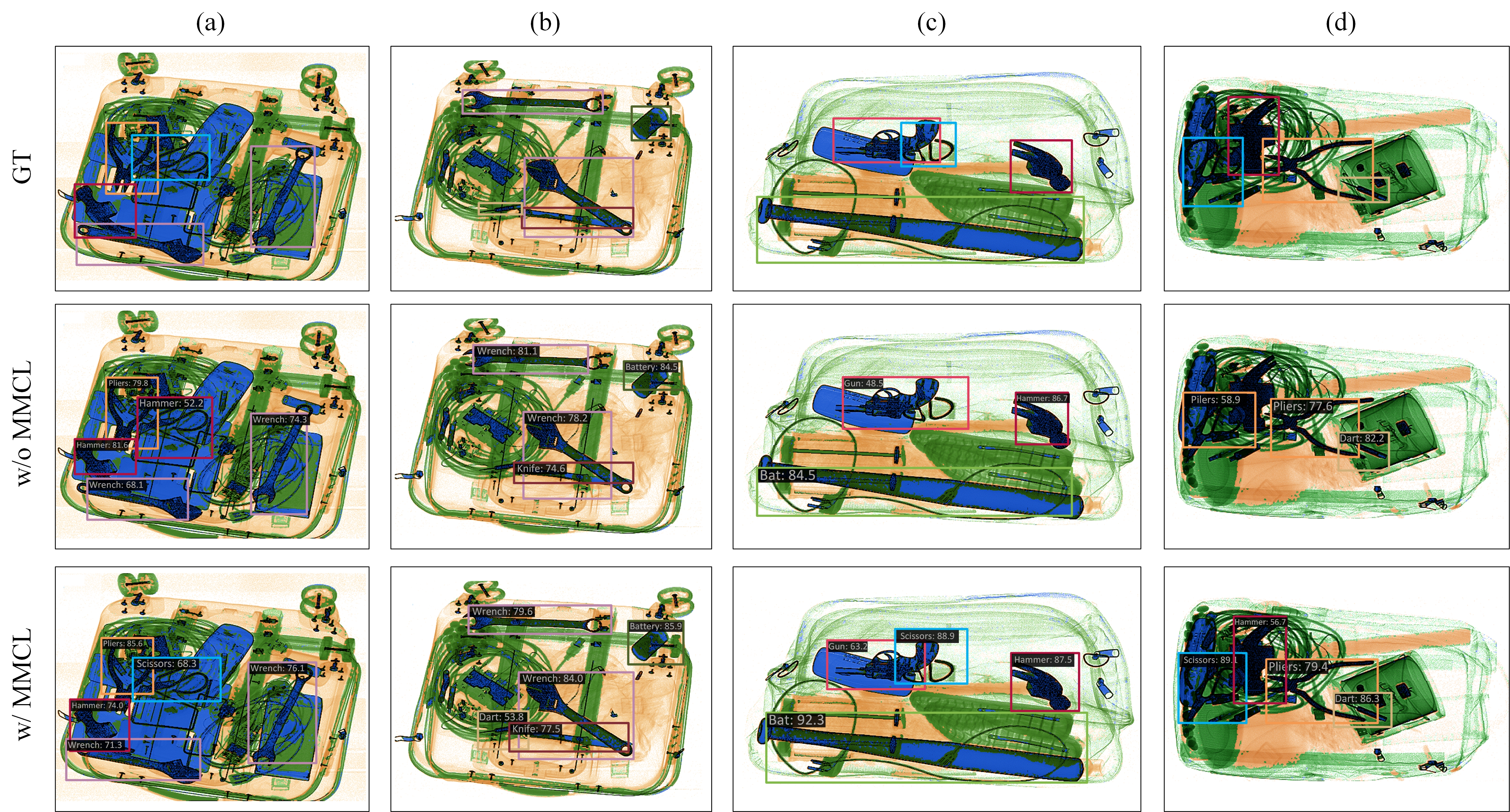}
    \caption{Visualization of prediction results from DINO~\cite{DINO} (middle row) and DINO+MMCL (bottom row) on the PIXray dataset. The top row presents the ground truth annotations. DINO exhibits misclassification or inaccurate localization under severe overlapping conditions, while the integration of MMCL enhances detection precision and consistency for prohibited items.}
    \label{fig:fig7}
\end{figure*}

\subsection{Analysis of Anti-Overlapping Ability}

\textbf{Qualitative Analysis}. \cref{fig:fig7} illustrates detection results on the PIXray~\cite{PIXray} dataset, comparing the performance of the DINO model before and after integrating MMCL. In specific challenging detection scenarios, the detector with MMCL more effectively identifies true positives (TP), such as successfully detecting darts in column (b) and scissors in column (c), indicating that MMCL can improve the model’s recall (TP/P). In more complex scenarios, the baseline model may miss some targets, producing false negatives (FN), and may also generate false positives (FP). For example, in column (a), the baseline model incorrectly identifies scissors as hammers. Similarly, in column (d), it fails to detect the hammer and misclassifies the scissors as pliers. In contrast, the model with MMCL achieves accurate classification and correctly predicts ground-truth boxes, resulting in a reduced miss rate (FN/P). Overall, these visualizations show that MMCL enhances the detection performance of DETR-like models on X-ray images with overlapping objects, thereby validating the effectiveness of the proposed method. 
\begin{figure}[!t]
\centering
\includegraphics[width=0.6\linewidth]{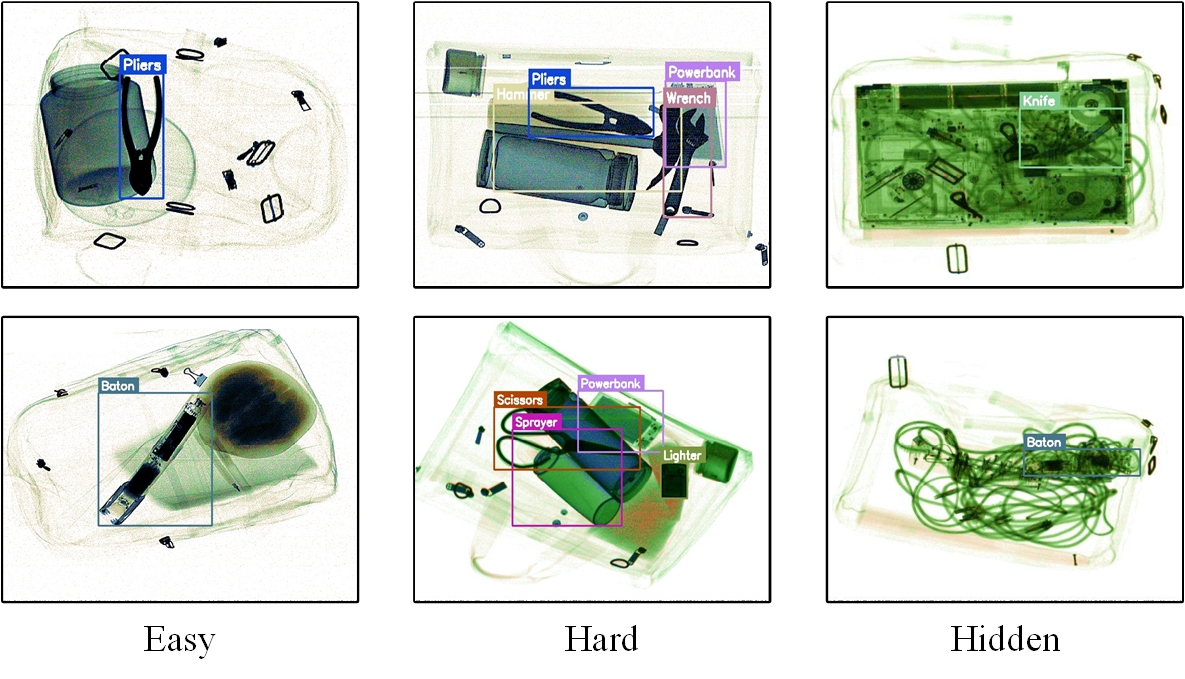}
\caption{Illustration of the three test sub-datasets in the PIDray dataset~\cite{PIDray}, designed to evaluate model robustness under varying degrees of object overlap. \textbf{Easy:} images with minimal or no overlap. \textbf{Hard:} prohibited items overlap with other objects or the background. \textbf{Hidden:} prohibited items are barely visible due to severe overlapping.}
\label{fig:fig8}
\end{figure}

\begin{table}[!ht]
\footnotesize
\begin{center}
\caption{Quantitative Analysis of MMCL's Anti-overlapping Ability.}
\label{tab:table10}
\begin{tabularx}{0.75\linewidth}{l|AAAA}
\toprule
Method  & AP$_{\text{Easy}}$ & AP$_{\text{Hard}}$ & AP$_{\text{Hidden}}$&AP$_{\text{Overall}}$ \\
\midrule
DINO &66.5&55.9 &41.7&54.7 \\
\rowcolor{cyan} DINO + MMCL &66.7 \bl{(+0.2)} &56.8 \bl{(+0.9)} &42.8 \bl{(+1.1)} &55.4\\
\bottomrule
\end{tabularx}
\end{center}
\end{table}

\textbf{Quantitative Analysis}.
We compare DINO with and without MMCL to evaluate the anti-overlapping capability conferred by MMCL on the three test sub-datasets of PIDray~\cite{PIDray}, namely Easy, Hard, and Hidden, which exhibit progressively more severe overlapping, as illustrated in~\cref{fig:fig8}. The quantitative results in~\Cref{tab:table10} show that MMCL is particularly effective for images with pronounced overlapping: the greater the overlap, the more substantial the accuracy improvement, with gains of +0.2\% AP$_\text{Easy}$, +0.9\% AP$_\text{Hard}$, and +1.1\% AP$_\text{Hidden}$.

\subsection{Computational Complexity}
As presented in~\Cref{tab:table11}, we analyze the computational overhead introduced by our MMCL mechanism using DINO and DINO+MMCL as examples. During training, the additional GPU memory requirement is minimal, increasing only from 1754 MB to 1759 MB, and the average iteration time per batch increases by merely 0.0038 s. Notably, during inference, MMCL does not participate in the prediction process, so neither the frame rate (FPS) nor the floating-point operations (FLOPs) are affected. Furthermore, the total number of parameters remains unchanged, as MMCL introduces no additional learnable parameters. These results demonstrate that MMCL is computationally lightweight during training and imposes no additional complexity during inference, making it an efficient enhancement for DETR-like models.

\begin{table}[!ht]
\footnotesize
\caption{Complexity analysis of the MMCL framework.}
\centering
\begin{tabularx}{0.9\linewidth}{l|A|A|A|A|A}
\toprule
\multirow{2}*{Method}&\multicolumn{2}{c|}{Training}&\multicolumn{2}{c|}{Inference}&\multirow{2}*{\#Params}\\
\cmidrule{2-5}
~&Memory& Time& FPS & \#GFLOPs &~  \\
\midrule
DINO& 1754M& 0.1340s&54 &26.89 &58.38M\\
\rowcolor{cyan} DINO + MMCL& 1759M& 0.1378s&54 &26.89 &58.38M\\
\bottomrule
\end{tabularx}
\label{tab:table11}
\end{table}

\subsection{Limitation} 
We also train DINO~\cite{DINO} with MMCL on COCO~\cite{COCO}, a natural-image object detection dataset, and observe that MMCL provides little to no improvement. This limited generalization arises from two key factors. First, overlapping objects in COCO often result in complete occlusion, whereas in X-ray security images, they remain partially visible, allowing detectors to extract prohibited items more reliably. Second, the physical size of class-specific objects in X-ray images is relatively stable due to the fixed imaging distance, while in natural images, object sizes vary significantly with distance and perspective. Consequently, features of class-specific objects in natural images are less consistent, reducing the effectiveness of partitioning content query distributions into clusters aligned with object classes. Each cluster thus struggles to learn a stable class-specific prior, limiting MMCL’s impact on natural image detection.
\section{Conclusion}\label{Conclusion}
This paper presents MMCL, a novel contrastive learning framework designed to enhance the anti-overlapping capability of DETR-like object detectors for X-ray images. By explicitly refining the content query distribution via contrastive supervision, MMCL promotes category-consistent query representations and mitigates semantic confusion arising from overlapping objects. 
Comprehensive experiments across different backbones, DETR variants, contrastive losses, and datasets, demonstrate that MMCL consistently improves detection accuracy across varying levels of overlap, with negligible computational overhead and no additional parameters during inference.
Beyond empirical improvements, MMCL offers conceptual insight into the importance of correcting content query distributions in transformer-based object detection. It shows that enforcing contrastive consistency among content queries effectively stabilizes category priors and enhances model generalizability. Although its performance on natural image datasets remains limited due to higher variability in occlusion and object scale, the framework presents a promising direction for X-ray image domains such as security inspection and medical diagnosis. Future work will explore adaptive query grouping and cross-domain representation learning to further extend MMCL’s applicability. Overall, MMCL represents a lightweight yet principled advancement toward more robust X-ray object detection, highlighting the potential of contrastive learning to address intrinsic challenges in overlapping object detection.
 




\Acknowledgements{This work is supported by the National Natural Science Foundation of China under Grant U22A2063, 62173083, 62276186, and 62206043; the China Postdoctoral Science Foundation under No.2023M730517 and 2024T170114; the Liaoning Provincial "Selecting the Best Candidates by Opening Competition Mechanism" Science and Technology Program under Grant 2023JH1/10400045; the scholarship of China Scholarship Council (202506080096); the Fundamental Research Funds for the Central Universities under Grant N2424022; the Major Program of National Natural Science Foundation of China (71790614) and the 111 Project (B16009).}



\bibliographystyle{scis}
\bibliography{main}






\end{document}